%% file: root.tex
\documentclass[graybox]{styles/svmult}

\usepackage{mathptmx}       
\usepackage{helvet}         
\usepackage{courier}        
\usepackage{type1cm}        
\usepackage{makeidx}         
\usepackage{graphicx}        
\usepackage{multicol}        
\usepackage[bottom]{footmisc}

\usepackage{amsmath}
\usepackage{amsfonts}
\usepackage{wrapfig, url}
\usepackage{siunitx}
\usepackage{hyphenat}

\usepackage{algpseudocode}
\usepackage{algorithm}

\makeindex             

\begin{document}

\title*{Meander Based River Coverage by an Autonomous Surface Vehicle}
\author{Nare Karapetyan, Jason Moulton, and Ioannis Rekleitis}
\institute{N. Karapetyan, J. Moulton, and I. Rekleitis \at University of South Carolina, Columbia, SC, \email{[nare,moulton]@email.sc.edu, yiannisr@cse.sc.edu}}
%
%
\maketitle


\abstract{Autonomous coverage has tremendous importance for environmental surveying and exploration tasks performed on rivers both in terms of efficiency and data collection quality. Most surveys of rivers are performed manually using quite similar approaches. Using these practices to automate these processes improves the quality of survey operations. In addition to human expertise on the type of patterns, the coverage of a river can be optimized using the river meanders to determine the direction of coverage. In this work we use the implicit information on the speed of the water current, inferred from the curves of the river, to reduce the cost of coverage. We use autonomous surface vehicles (ASVs) to deploy the proposed methods and demonstrate the efficiency of our method. In addition we compare the proposed method with previous coverage techniques developed in our lab. When taking into account meanders the coverage time has been decreased in average by more than $20\%$. The deployments of the ASVs were performed on the Congaree River, SC, USA, and resulted in more than 27 km of total coverage trajectories. }

\input{intro}

\input{relatedWork}
\input{proposed_method}
\input{results}
\input{conclusion}
\input{acknowledgment}

\bibliographystyle{styles/spmpsci}
\bibliography{root}

\end{document}

%% file: intro.tex
\section{Introduction}
\label{sec:Intro}

Significant work has been done in the field of autonomous area coverage ~\cite{galceran2013survey, Choset-2001}, utilizing single or multiple robot systems. Depending on the goal or constraints of the coverage operation different variations have been presented: coverage under limited resources ~\cite{sipahioglu2010energy, lewis2017semi}, information driven coverage \cite{manjanna2016efficient}, complete coverage \cite{karapetyan2017efficient, RekleitisAMAI2008b}, sampling based coverage, passive coverage \cite{kwok2010deployment}. One of the most common approaches is known as boustrophedon coverage, that performs back and forth or “lawn mower” motions~\cite{Acar2002J1,Acar2002J2}. This method performs very well in open areas that do not have many changes in the shape~\cite{lewis2017semi,RekleitisAMAI2008b}. Coverage is a central problem in multiple domains, such as: agriculture, environmental monitoring, search and rescue, and marine exploration. For each domain the desired solution has different priorities, and can be influenced by different factors. Therefore, having  prior knowledge about the physics of the environment can enhance coverage strategies applied in that domain. 

\begin{wrapfigure}[12]{r}{0.5\textwidth}
\centering
\vspace{-0.2in}
\includegraphics[width=0.5\textwidth]{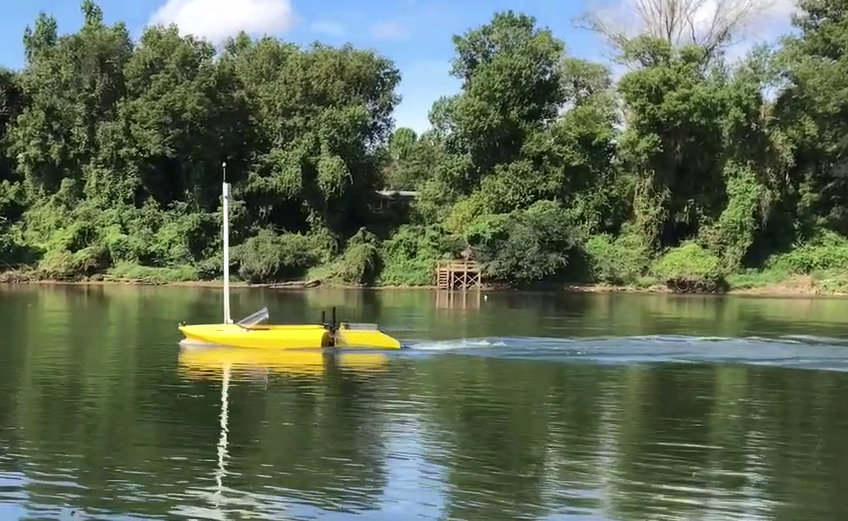}
\label{fig:beauty}
\vspace{-0.2in}
\caption{Autonomous Surface Vehicle (ASV) during coverage operations at the Congaree River.}
\end{wrapfigure}
To the best of our knowledge we were the first to implement systematic coverage in a riverine environment~\cite{karapetyan2019riverine}. These algorithms differ from well known approaches in a way that they provide domain specific geometric solutions. For the tight and uneven spaces prior algorithms were not as effective. The following work addresses the river coverage problem, using current flow speed information conditioned by the meanders of the river, to enhance time and energy efficiency of the robot’s operation.

River morphology is quite complicated, especially, the twists and turns a river takes, called meanders. Due to these curves, the water flow at different speeds across the river, which in turns contributes to different rates of erosion and sediment deposits. ASV operations are heavily affected by the water currents encountered~\cite{MoultonISER2018}, as such, the efficiency of the ASV's navigation can be improved by choosing to go against the slower currents and with the faster currents.  

In the following section we present a survey of related work. Next, Section~\ref{sec:method} will define the coverage problem and present the meander based algorithm with some discussion on improvement of that method as well.  Section~\ref{sec:res} presents the experimental setting and the results of field trials with discussion of outcomes. Finally Section~\ref{sec:concl}, gives a summary of the proposed method and some remarks on the future work.

%% file: relatedWork.tex
\section{Related Work}
\label{sec:related}

Substantial work has been done on the design and operations of autonomous surface vehicles (ASVs) in rivers. One of the works show how to design and operate ASVs for performing bathymetric surveys~\cite{ferreira2009autonomous}. There has been some work also studying the problem of navigating a river with an ASV \cite{snyder2004autonomous}.

In our recent work on river coverage~\cite{karapetyan2019riverine} we have proposed different coverage techniques that vary by the available resources and the application domain. One of the approaches proposed was termed zig\hyp zag (Z\hyp cover) approach performing a single pass over a river segment by bouncing back and forth across the shores of the river. Such an approach is utilized when there is limited time available (e.g. fuel constraints). Another approach termed Transversal Coverage (T-cover) performed boustrophedon coverage in between the shores. While inefficient due to the many turns, T\hyp cover is utilized during sediment studies. Finally, the most efficient was Longitudinal coverage (L-cover), which is a complete coverage algorithm. It decomposes the river along the length based on the width information, such that each cluster of the river has approximately the same width. Then each of the clusters is decomposed into passes based on the size of the sensor footprint of the vehicle. Nevertheless in that work we do not consider any of the geological properties of rivers. As such some work has been done to find connection between river meanders and the speed of river current ~\cite{qin2017robots, njenga2012velocity} or to model the external forces ~\cite{MoultonISER2018}. As a matter of fact the first one that understood how flow affects the length of meanders and its down-flow migration was Albert Einstein~\cite{einstein1926cause}. In another work Kai Qin and Dylan A. Shell use the well studied model of the geometry of meanders to estimate the shape of the unseen portion of the river~\cite{qin2017robots}. Using this information as an input, a boat can adjust the speed and perform more optimal and smooth paths when performing online navigation.

The problem approached in this paper is a variation of the well-studied coverage problem \cite{galceran2013survey}. Of particular relevance are two works dealing with the coverage of rivers using drifters: vehicles that do not have sufficient power to travel against the current \cite{kwok2010deployment,kwok2010coverage} and another work dealing with coverage path planning for a group of energy-constrained robots \cite{sipahioglu2010energy}. One notable work breaks from the tendency to emphasize complete coverage, instead attempting to conserve time and fuel by focusing coverage on regions of interest \cite{manjanna2016efficient}. This allowed them to create a map of a coral reef area with half the distance coverage and power use than a lawnmower-style complete coverage algorithm would have required. Another paper, in which lawnmower-style coverage is applied to a Dubins vehicle, reformulates the problem as a variant of the Traveling Salesman Problem in order to obtain an optimal solution~\cite{lewis2017semi}.


Even though, there is substantial work done in the literature for area coverage, they do not address the coverage in tight and uneven spaces, such as rivers. Moreover, to the best of our knowledge no one has addressed the efficiency problem of river navigation taking into account meanders information or any other river specific variables. Meanwhile autonomous navigation on river can help reduce long and expensive exploration expeditions that scientists have to take for studying or monitoring rivers or marine environments in general. 

In this work, we address the autonomous coverage problem for river surveying by taking advantage of the river meanders. We use the implicit information encoded in the shape of meanders regarding the speed of the river to choose a more efficient route when travelling downstream or upstream. In the following section, the problem is formally defined and the meander based coverage approach is described with a suggestion for a possible improvement.

%% file: proposed_method.tex
\section{Proposed Method}
\label{sec:method}

The objective of this work is to perform complete coverage of a river segment by taking into account  the varying speeds of water current across the river. We study the problem with an Autonomous Surface Vehicle (ASV) that is deployed on a known environment. The ASV is equipped with different types of depth sensors. We acquire the environment map from Google satellite images and convert it to an occupancy grid map $M: \mathbb{R}^2 \to \{0, 1\}$.  Values of 0 indicate the portion of the river that is within the region of interest, while 1 indicates locations that we treat as obstacles. Assuming that the exact bounding box of the interest region is given together with starting point $v_s$, we can infer implicitly the general direction of the coverage.

\subsection{Meander Based Coverage (M-cover)}
\label{subsec:m-cover}

\begin{figure*}
\sidecaption
\begin{tabular}{cc}
\includegraphics[width=0.48\textwidth]{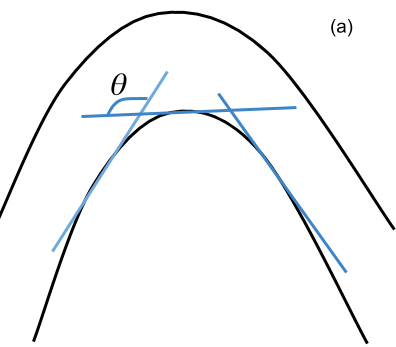}
\label{fig:expl1}&
\includegraphics[width=0.5\textwidth]{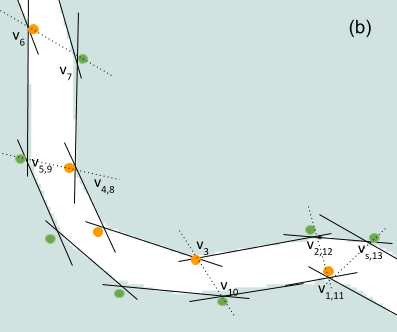}
\label{fig:res}
\end{tabular}
\caption{A sketch for finding the meanders. (a) The procedure of checking the intersection of a neighbor pair of tangents. (b) The order of vertices the algorithm will visit if coverage is to be performed upwards. }
\label{fig:explained}
\end{figure*}

In the meander based coverage we are assuming that on the inner bend, the downriver speed of the current is slower from any neighbour region closer to the outside bend of the river. As it is shown in Figure \ref{fig:explained} (b) along the passes that connect green dots the water flow is faster, whereas orange ones indicate region where the flow is slower. To find the meanders, we are looking into the intersection of two consecutive tangent lines to the curve of the river contour (Figure \ref{fig:explained} (a)). If the lines intersect inside the river then inner curve is identified (orange vertex), otherwise if the intersection is on land then the outside bend is found (green vertex). Using this information the M-cover algorithm depicted in Algorithm~\ref{alg:mcover} finds an efficient complete coverage path. It takes as input the map of the river $M$, a starting point $v_s$, and the spacing width information (sensor footprint size). First, the direction of the coverage is identified implicitly from $v_s$ and $M$, then, the directional contour $C_{vec}$ is generated (Line 2). Consequently, the river is split into $S_{vec}$ segments, utilizing the  meander information, using the above explained intuition (Line 4). Each segment is split into segments that the robot can cover in a single pass (Line 6). We decompose area into even number of segments in order to return back to the initial starting point $v_s$. Each of the passes are assigned a direction: the first pass that is closest to the inner bank of the river is getting reserved for upwards travel, whereas the ones closer to the outer side are getting reserved for downwards travel (Lines 7-14). A pass is added between each consecutive segment of meanders: from orange to the closest green on the opposite edge of the river (Lines 16-17).

\begin{algorithm}
\caption{M-Cover}
\label{alg:mcover}
\textbf{Input:} binary map of river $M$, starting point\ $v_s$, spacing parameter $s$\\
	\textbf{Output:} a $\pi$ path
	\begin{algorithmic}[1]
\State $\textit{$\Delta$w} \gets \textrm{initialize}()$
\State $\textit{$C_{vec}$} \gets \textrm{getDirectionalContours}(M)$
\State $\textit{$\alpha$} \gets \textrm{getDownRiverDirection}(\textit{$C_{vec}$}, \textit{$v_s$})$
\State $S_{vec} \gets\textrm{getMeanderSegments}(\Delta w, \textit{$C_{vec}$}, \alpha)$
\For{\textbf{each} $S \in S_{vec}$}
\State $P \gets splitIntoEvenPasses(S, C_{vec}, s)$
\State $k \gets |P|$
\For{\textbf{each} $p_i, p_{k/2+i} \in P$}
\If{$p_i~is~on~outside~bend$}
\State{$p_i \gets ~down~direction, p_{k/2+i} \gets ~up~direction$}
\Else
\State{$p_{k/2+i} \gets ~down~direction, p_{i} \gets ~up~direction$}
\State append $p_i$, $p_{k/2+i}$ to $\pi$
\EndIf
\EndFor
\State{$S_{prev} \gets S$}
\State{$p \gets createPassBetween(S_{prev}, S)$}
\State append $p$ to $\pi$
\EndFor
\State \Return $\pi$
\end{algorithmic}
\end{algorithm}

\begin{figure*}
\sidecaption
\begin{tabular}{cc}
\includegraphics[width=0.5\textwidth]{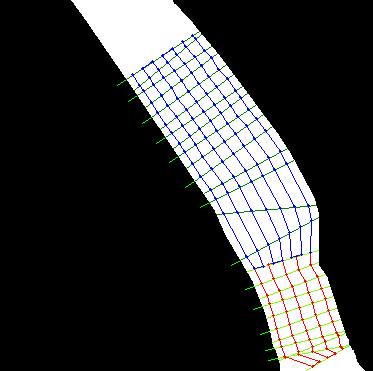}
\label{fig:expl}&
\includegraphics[width=0.5\textwidth]{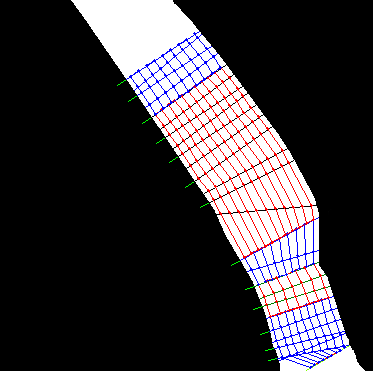}
\label{fig:res1}
\end{tabular}
\caption{An example of splitting an area into segments based on the width information. }
\label{fig:cluster}
\end{figure*}

The simple M-cover approach does not take into account the change in the width of the river, which can affect the number of the passes one can generate. To solve that problem we propose to adapt the L-cover algorithm introduced in previous work on river coverage~\cite{karapetyan2019riverine}. With this modification the algorithm will perform coverage in segments that have the same width (see Algorithm~\ref{alg:lm-cover}). In the same way the width based approach will take as an input the map $M$, the starting point $v_s$ and the spacing information. In this case we simply apply the L-Cover algorithm, to split the area into regions that have approximately the same width (Line 4), and then on each of those segments we apply the M-Cover algorithm to generate the more efficient path. The Figure~\ref{fig:cluster} shows the resulting clusters for the same segment of river with different spacing value $s$.

\begin{algorithm}
\caption{Width Based M-Cover}
\label{alg:lm-cover}
\textbf{Input:} binary map of river $M$, starting point\ $v_s$ spacing parameter $s$\\
	\textbf{Output:} a $\pi$ path
	\begin{algorithmic}[1]
\State $\textit{$C_{vec}$} \gets \textrm{getDirectionalContours}(M)$
\State $\textit{$\theta$} \gets \textrm{getDownRiverDirection}(\textit{$C_{vec}$}, \textit{$v_s$})$
\State $Cl_{vec} \gets getSameWidthClusters({C_{vec}, \theta, s})$
\For{\textbf{each} $Cl \in Cl_{vec}$}
\State $\textit{p} \gets \textrm{M-Cover}(\textit{Cl}, v_s, \textit{s})$
\State append $p$ to $\pi$
\EndFor
\State \Return $\pi$
\end{algorithmic}
\end{algorithm}

%% file: results.tex
\section{Results}
\label{sec:res}

\begin{figure*}
\sidecaption
\begin{tabular}{cc}
\includegraphics[width=0.6\textwidth]{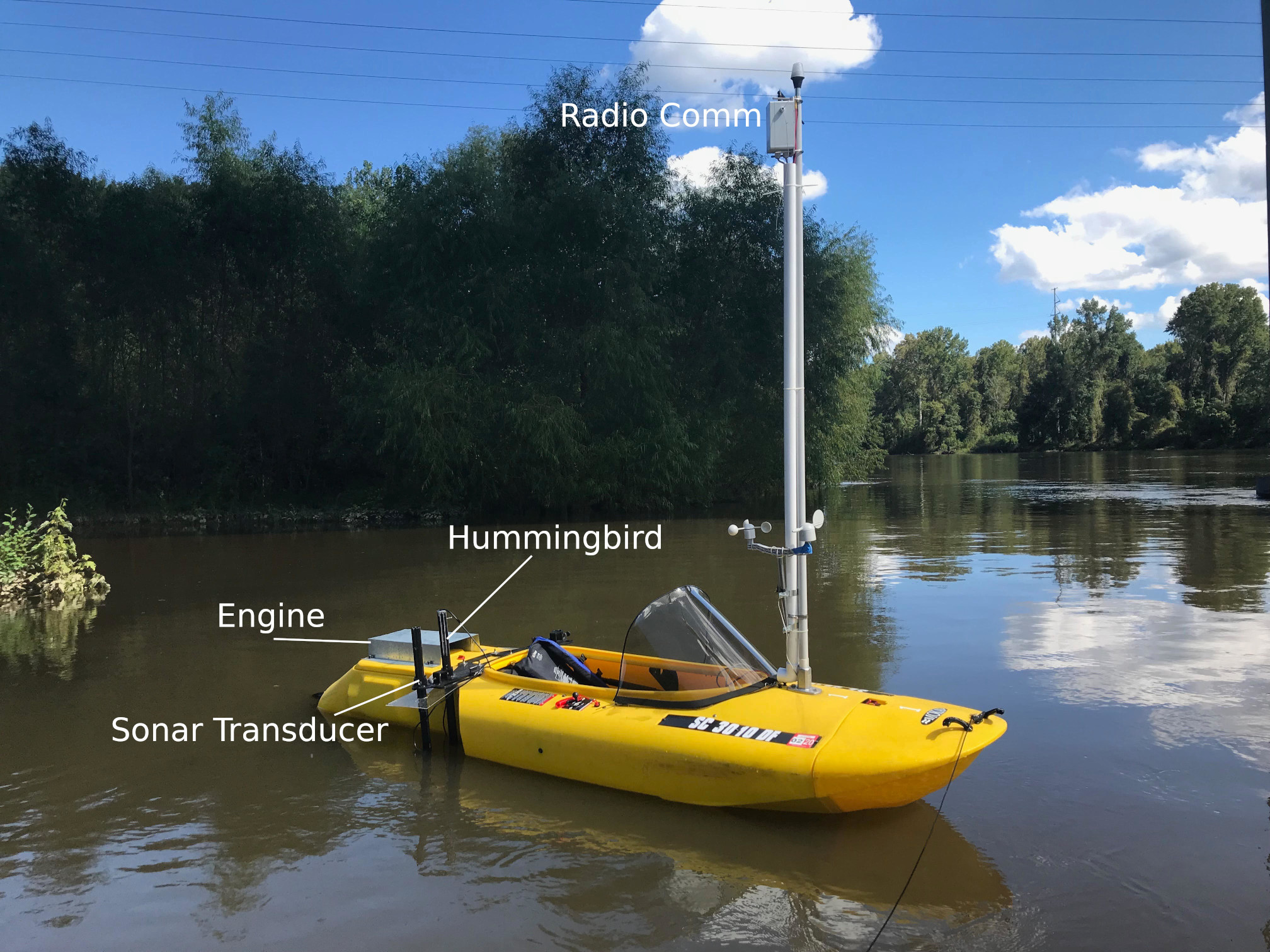}
\label{fig:r1}
\end{tabular}
\caption{The AFRL jetyak used during the field deployments. The main controller is the PixHawk that performs GPS-based way-point navigation. In addition different depth sensors are mounted on the boat for surveying operations.}
\label{fig:jet}
\end{figure*}

The proposed M-cover algorithm was deployed with an ASV on different sizes of river segments, performing overall nearly 27km of coverage on the Congaree River,SC. We have used the AFRL jetyaks \cite{MoultonOceans2018} equipped with a PixHawk controller for performing GPS-based waypoint navigation, a Raspberry Pi computer which runs the Robot Operating System (ROS) framework~\cite{ros} and records sensor data and GPS coordinates (Figure ~\ref{fig:jet}).

\begin{figure*}
\sidecaption
\begin{tabular}{cc}
\includegraphics[width=0.455\textwidth]{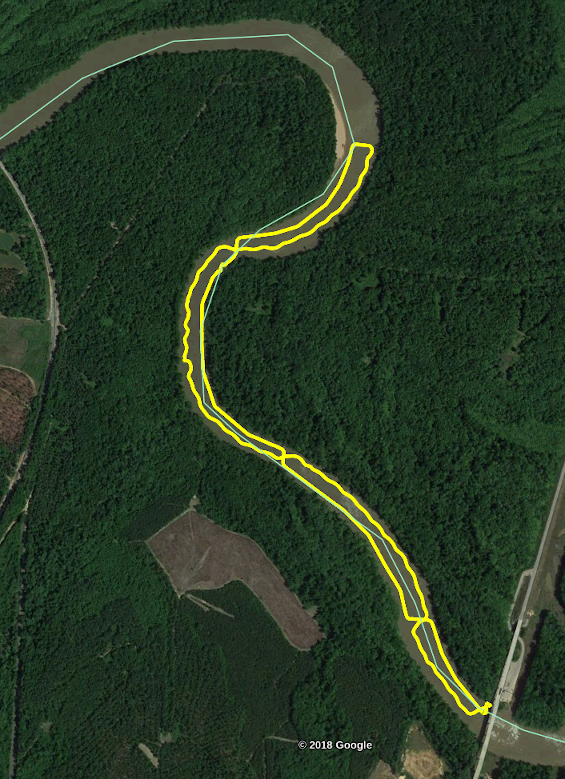}
\label{fig:r11}&
\includegraphics[width=0.5\textwidth]{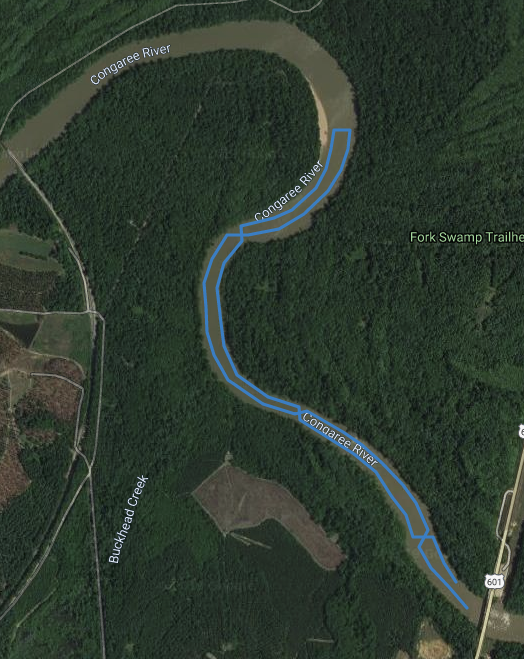}
\label{fig:r2}
\end{tabular}
\caption{The way-points generated by the M-cover algorithm (right) with the actual GPS trajectory (left) executed on the 2.76km long river segment. }
\label{fig:m-cover-short}
\end{figure*}

\begin{figure*}
\sidecaption
\begin{tabular}{cc}
\includegraphics[width=0.455\textwidth]{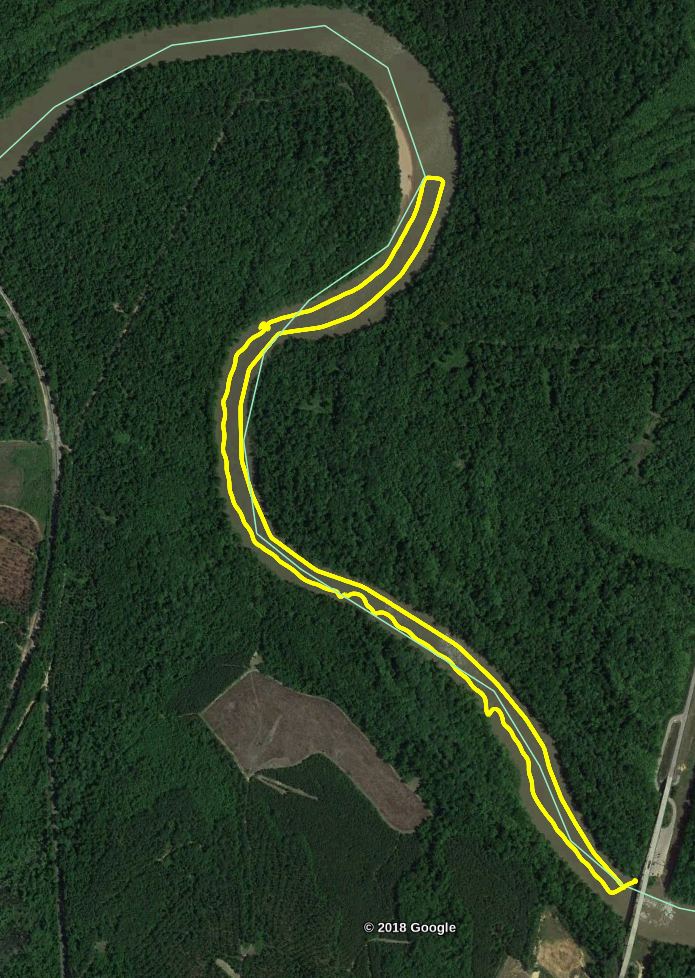}
\label{fig:r3}&
\includegraphics[width=0.5\textwidth]{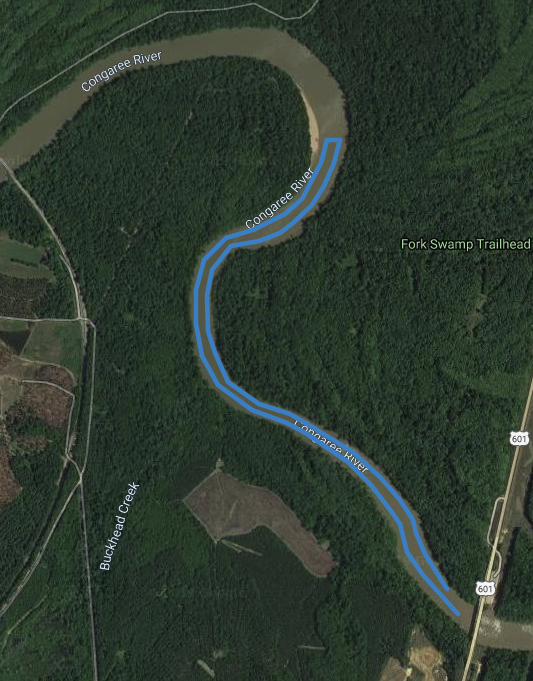}
\label{fig:r4}
\end{tabular}
\caption{The way-points for covering a river segment without taking into account meanders (right) with the actual GPS trajectory (left) executed on the 2.76km long river segment. }
\label{fig:l-cover-short}
\end{figure*}

\begin{figure*}
\sidecaption
\begin{tabular}{cc}
\includegraphics[width=0.5\textwidth]{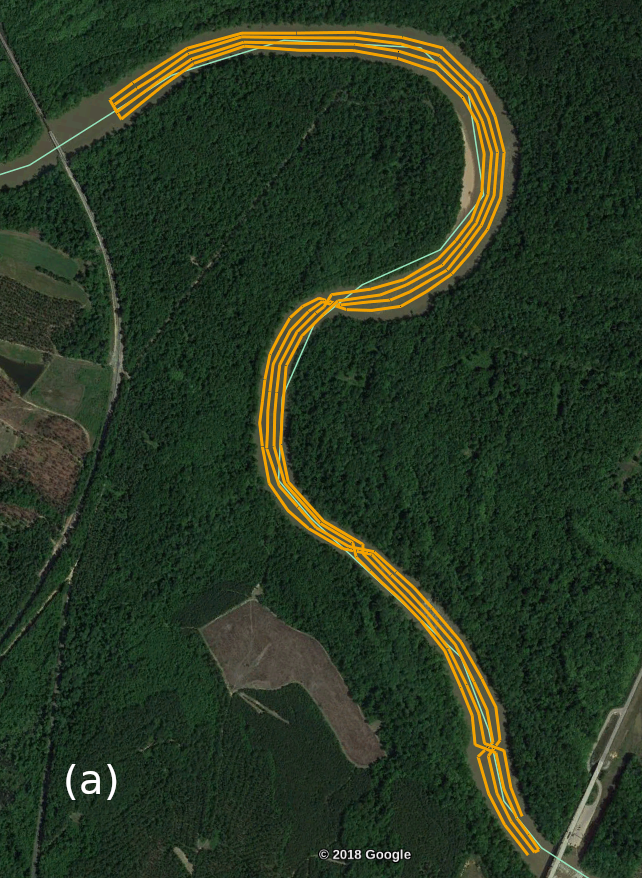}
\label{fig:m1}&
\includegraphics[width=0.5\textwidth]{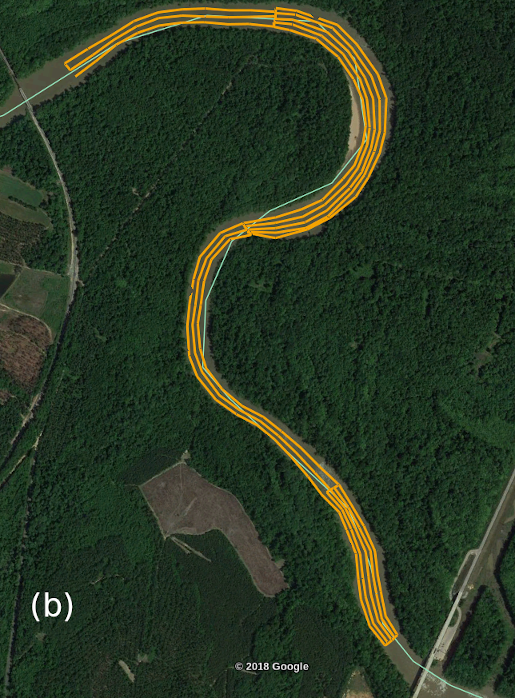}
\label{fig:l1}\\
\includegraphics[width=0.5\textwidth]{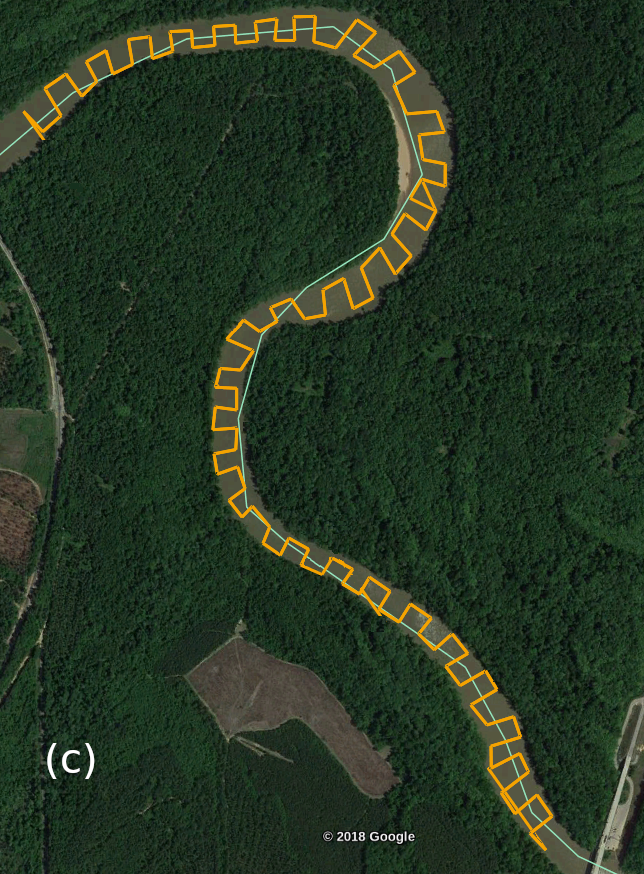}
\label{fig:p1}&
\includegraphics[width=0.5\textwidth]{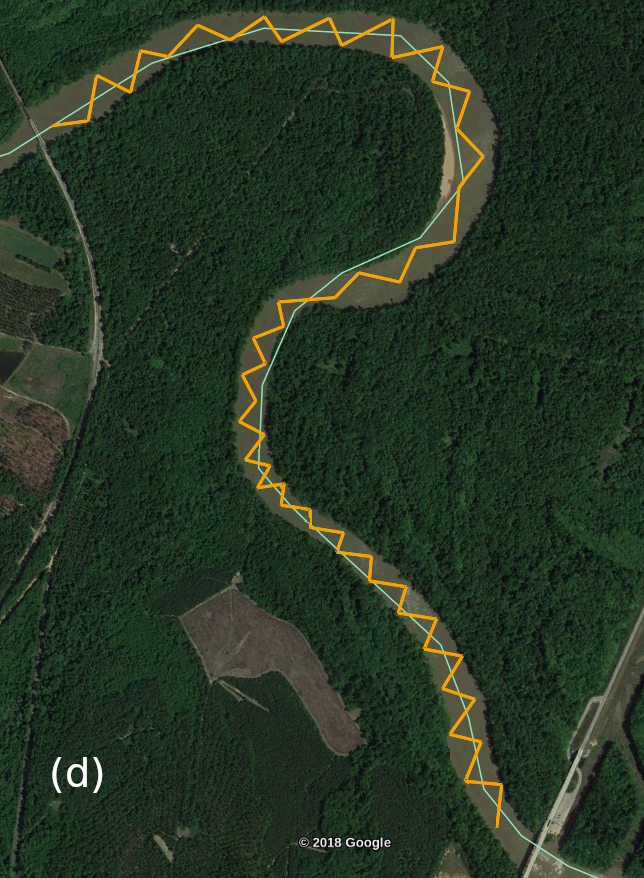}
\label{fig:t1}
\end{tabular}
\caption{Way-points of the different coverage patterns executed on the 4.2km long segment of Congaree river, SC. (a) M-cover, (b) L-cover, (c) T-cover, (d) Z-cover} 
\label{fig:all}
\end{figure*}

The main objective of the experiments is to demonstrate that even with dynamically changing environments the proposed M-cover approach ensures more efficient coverage. We have deployed the ASV on 4.12 km and 2.76 km segments of the Congaree river; see Table~\ref{tab:1}. The width of the river in average is 90m. The long segment was covered with small value of spacing which resulted in four passes (Figure ~\ref{fig:all} (a)), whereas in the smaller segment the ASV performed only two passes (Figure ~\ref{fig:m-cover-short}). In addition, we run the L-cover algorithm with same spacing for the smaller region and similar to the M-cover two passes were generated (Figure ~\ref{fig:l-cover-short}). For the longer region with the same spacing value the L-cover generated segments with either three or five passes, though it resulted with a similar length of the coverage trajectory (Figure ~\ref{fig:all} (b)). When the execution time of the coverage operation of M-cover is compared with the  L-cover for both experiments the M-cover is  on average $20\%$ more efficient. Note that results are only based on the performed field trials. It has been observed that the river current data change even in an hourly base. Therefore, generating a graph model that will represent the approximate currents would not be a comprehensive representation of a real world scenario.

In addition we have sampled small portions of the river and compared the coverage time for the two opposite banks of the river to show the affect of the current on coverage time. The results showed that when travelling upstream on the outside portion of the meander the coverage time is almost twice longer than if going downstream (approximately $47\%$).

In Figure \ref{fig:all} we show the way-points generated by the different river coverage algorithms to illustrate the different patterns. If looking into complete coverage algorithms we have shown in our previous work that L-cover is the more efficient approach, thus we compared the execution times of proposed approach only with the L-cover approach. The T-cover approach fails mostly because of the cost of travelling back to the starting position, whereas the M-cover still performs coverage on the way back.

\begin{table}
\caption{The experimental results of deployments}
\label{tab:1}       
%
%

\begin{tabular}{p{2.5cm}p{2.4cm}p{2.5cm}p{2cm}p{1.8cm}}
\hline\noalign{\smallskip}
Algorithm & Total Area & Path Length & Duration & \# of Passes  \\
\noalign{\smallskip}\svhline\noalign{\smallskip}
M-Cover & 4.12km x 90m  & 16.6km & 2h 55m & 4\\
L-Cover & 4.12km x 90m   & 16.3km & 3h 35m & 3 and 5 \\
M-Cover &  2.76km x 90m & 5,32km & 47m 36s & 2\\
L-cover & 2.76km x 90m  &  5.13km & 59m 46s & 2\\
\noalign{\smallskip}\hline\noalign{\smallskip}
\end{tabular}
\end{table}

\begin{figure*}
\sidecaption
\begin{tabular}{cc}
\includegraphics[width=0.5\textwidth]{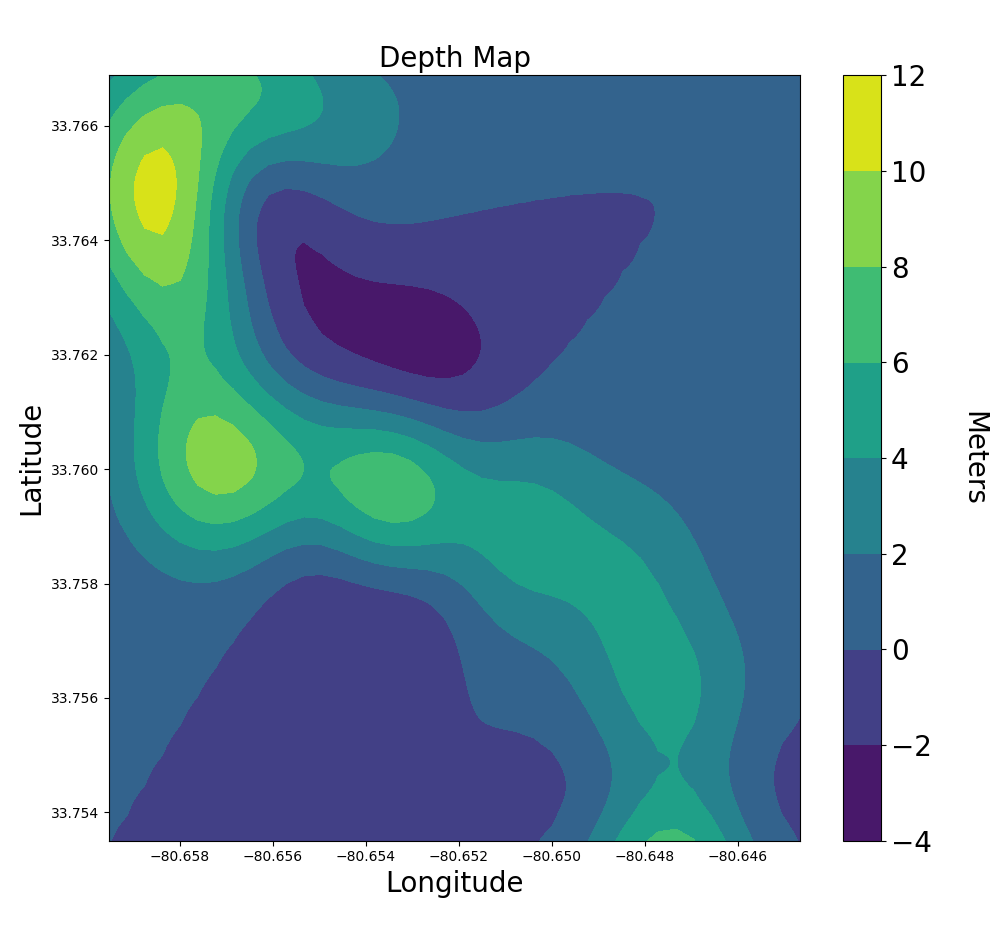}
\label{fig:m11}&
\includegraphics[width=0.5\textwidth]{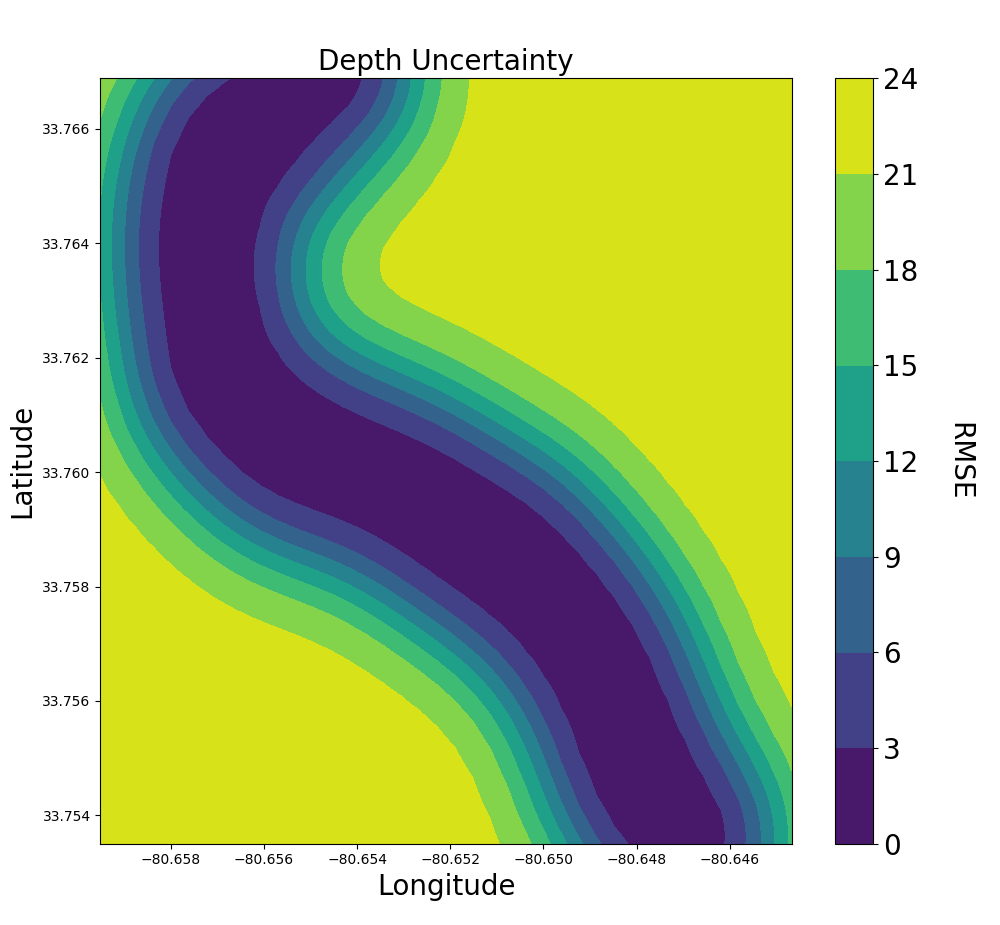}
\label{fig:l11}
\end{tabular}
\caption{Depth map generated by Gaussian Process with RMSE on the left, indicating the accuracy of the prediction.}
\label{fig:depth}
\end{figure*}

In addition from the data collected on the smaller river segment with a CruzPro  DSP  Active  Depth,  Temperature  single ping  SONAR  Transducer, we have generated the depth map of the river segment using a Gaussian Process method (see Figure ~\ref{fig:depth}). 

%% file: conclusion.tex
\section{Conclusion}
\label{sec:concl}

The following work introduced a new coverage method specifically designed for rivers. It takes into account the connection of the bend of curve on meanders to choose the most optimal way of travelling along the river. The proposed M-cover method is a complete and more efficient coverage approach. We have demonstrated the improvements of the coverage performance by experimental results. The experiments were performed with an ASV on 4.12km and 2.76km long segments of the Congaree River, SC. In our previous work on river coverage we showed  that a complete coverage method termed L-cover is the most efficient. Therefore, we have deployed the ASV performing both the L-cover and the M-cover trajectories. The results demonstrated that the M-cover algorithm outperforms the L-cover approach by decreasing coverage time on average by $20\%$. In addition, we proposed a possible improvement of the M-cover by combining it with the L-cover algorithm to take into account also the width of the river, which can optimize also the number of passes generated per section. Finally, the depth map was generated using the data collected from CruzPro DSP single ping SONAR Transducer.

Taking into account the challenges encountered during field deployments, obstacle avoidance strategies must be implemented for both underwater and above water obstacles. The important next contribution will be to extend this work on multiple robots~\cite{karapetyan2017efficient, karapetyan2018multi}. In addition given the dynamic changes of the environment, we are interested in generating a robust model of the water current of the river~\cite{MoultonISER2018}. With this model path planning will associate different cost values and perform adaptive coverage.

%% file: acknowledgment.tex
\section{Acknowledgment}
\label{sec:akl}

This work was partially supported by a SPARC Graduate Research Grant from the Office of the Vice President for Research at the University of South Carolina. The authors also would like to thank the National Science Foundation for its support (NSF 1513203).

%% file: root.bbl
\newcommand{\noop}[1]{}
\begin{thebibliography}{10}
\providecommand{\url}[1]{{#1}}
\providecommand{\urlprefix}{URL }
\expandafter\ifx\csname urlstyle\endcsname\relax
  \providecommand{\doi}[1]{DOI~\discretionary{}{}{}#1}\else
  \providecommand{\doi}{DOI~\discretionary{}{}{}\begingroup
  \urlstyle{rm}\Url}\fi

\bibitem{Acar2002J2}
Acar, E.U., Choset, H.: {Sensor-based Coverage of Unknown Environments:
  Incremental Construction of Morse Decompositions} \textbf{21(4)}, 345--366
  (2002)

\bibitem{Acar2002J1}
Acar, E.U., Choset, H., Rizzi, A.A., Atkar, P.N., Hull, D.: {Morse
  Decompositions for Coverage Tasks}.
\newblock The International Journal of Robotics Research \textbf{21(4)},
  331--344 (2002)

\bibitem{Choset-2001}
Choset, H.: Coverage for robotics – a survey of recent results
  \textbf{31(1-4)}, 113--126 (2001)

\bibitem{einstein1926cause}
Einstein, A.: The cause of the formation of meanders in the courses of rivers
  and of the so-called baer’s law.
\newblock Die Naturwissenschaften \textbf{14}(11), 223--224 (1926)

\bibitem{ferreira2009autonomous}
Ferreira, H., Almeida, C., Martins, A., Almeida, J., Dias, N., Dias, A., Silva,
  E.: {Autonomous bathymetry for risk assessment with ROAZ robotic surface
  vehicle}.
\newblock In: Oceans 2009-Europe, pp. 1--6. IEEE (2009)

\bibitem{galceran2013survey}
Galceran, E., Carreras, M.: A survey on coverage path planning for robotics.
\newblock Robotics and Autonomous systems \textbf{61}(12), 1258--1276 (2013)

\bibitem{karapetyan2017efficient}
Karapetyan, N., Benson, K., McKinney, C., Taslakian, P., Rekleitis, I.:
  Efficient multi-robot coverage of a known environment.
\newblock In: IEEE/RSJ International Conference on Intelligent Robots and
  Systems (IROS), pp. 1846--1852 (2017)

\bibitem{karapetyan2019riverine}
Karapetyan, N., Braude, A., Moulton, J., Burstein, J.A., Whitea, S., O'Kane,
  J.M., Rekleitis, I.: Riverine coverage with an autonomous surface vehicle
  over known environments.
\newblock In: IEEE/RSJ International Conference on Intelligent Robots and
  Systems (IROS) (2019)

\bibitem{karapetyan2018multi}
Karapetyan, N., Moulton, J., Lewis, J.S., {Quattrini Li}, A., O'Kane, J.M.,
  Rekleitis, I.M.: Multi-robot dubins coverage with autonomous surface
  vehicles.
\newblock In: 2018 {IEEE} International Conference on Robotics and Automation,
  {ICRA} 2018, Brisbane, Australia, May 21-25, 2018, pp. 2373--2379 (2018)

\bibitem{kwok2010coverage}
Kwok, A., Mart{\'\i}nez, S.: A coverage algorithm for drifters in a river
  environment.
\newblock In: American Control Conference (ACC), 2010, pp. 6436--6441. IEEE
  (2010)

\bibitem{kwok2010deployment}
Kwok, A., Mart{\'\i}nez, S.: Deployment of drifters in a piecewise-constant
  flow environment.
\newblock In: 49th IEEE Conference on Decision and Control (CDC), pp.
  6584--6589. IEEE (2010)

\bibitem{lewis2017semi}
Lewis, J.S., Edwards, W., Benson, K., Rekleitis, I., O'Kane, J.M.:
  Semi-boustrophedon coverage with a dubins vehicle.
\newblock In: 2017 IEEE/RSJ International Conference on Intelligent Robots and
  Systems (IROS), pp. 5630--5637. IEEE (2017)

\bibitem{manjanna2016efficient}
Manjanna, S., Kakodkar, N., Meghjani, M., Dudek, G.: {Efficient terrain driven
  coral coverage using Gaussian Processes for mosaic synthesis}.
\newblock In: Computer and Robot Vision (CRV), 2016 13th Conference on, pp.
  448--455. IEEE (2016)

\bibitem{MoultonOceans2018}
Moulton, J., Karapetyan, N., Bukhsbaum, S., McKinney, C., Malebary, S.,
  Sophocleous, G., {Quattrini Li}, A., Rekleitis, I.: An autonomous surface
  vehicle for long term operations.
\newblock In: MTS/IEEE OCEANS, Charleston, pp. 1--6 (2018)

\bibitem{MoultonISER2018}
Moulton, J., Karapetyan, N., {Quattrini Li}, A., Rekleitis, I.: {External Force
  Field Modeling for Autonomous Surface Vehicles}.
\newblock In: International Symposium on Experimental Robotics (ISER). Buenos
  Aires, Argentina (2018)

\bibitem{njenga2012velocity}
Njenga, K.J., Kwanza, J., Gathia, P.W.: Velocity distributions and meander
  formation of river channels.
\newblock International Journal of Applied \textbf{2}(9) (2012)

\bibitem{qin2017robots}
Qin, K., Shell, D.A.: Robots going round the bend—a comparative study of
  estimators for anticipating river meanders.
\newblock In: IEEE International Conference on Robotics and Automation (ICRA),
  pp. 4934--4940. IEEE (2017)

\bibitem{ros}
Quigley, M., Conley, K., Gerkey, B.P., Faust, J., Foote, T., Leibs, J.,
  Wheeler, R., Ng, A.Y.: {ROS: an open-source Robot Operating System}.
\newblock In: ICRA Workshop on Open Source Software (2009)

\bibitem{RekleitisAMAI2008b}
Rekleitis, I., New, A.P., Rankin, E.S., Choset, H.: Efficient boustrophedon
  multi-robot coverage: an algorithmic approach.
\newblock Annals of Mathematics and AI \textbf{52(2-4)}, 109--142 (2008)

\bibitem{sipahioglu2010energy}
Sipahioglu, A., Kirlik, G., Parlaktuna, O., Yazici, A.: Energy constrained
  multi-robot sensor-based coverage path planning using capacitated arc routing
  approach.
\newblock Robotics and Autonomous Systems \textbf{58}(5), 529--538 (2010)

\bibitem{snyder2004autonomous}
Snyder, F.D., Morris, D.D., Haley, P.H., Collins, R.T., Okerholm, A.M.:
  Autonomous river navigation.
\newblock In: Mobile robots XVII, vol. 5609, pp. 221--233. International
  Society for Optics and Photonics (2004)

\end{thebibliography}
